# Localization of Autonomous Vehicles: Proof of Concept for A Computer Vision Approach

**Sara Zahedian[1], Kaveh Farokhi Sadabadi[2], Amir Nohekhan[3]**
Center for Advanced Transportation Technology, University of Maryland, College Park, Maryland, USA

**Abstract**

This paper introduces a visual-based localization method for autonomous vehicles (AVs) that operate in the absence of any complicated hardware system but a single camera. Visual localization refers to techniques that aim to find the location of an object based on visual information of its surrounding area. The problem of localization has been of interest for many years. However, visual localization is a relatively new subject in the literature of transportation. Moreover, the inevitable application of this type of localization in the context of autonomous vehicles demands special attention from the transportation community to this problem. This study proposes a two-step localization method that requires a database of geotagged images and a camera mounted on a vehicle that can take pictures while the car is moving. The first step which is image retrieval uses SIFT local feature descriptor to find an initial location for the vehicle using image matching. The next step is to utilize the Kalman filter to estimate a more accurate location for the vehicle as it is moving. All stages of the introduced method are implemented as a complete system using different Python libraries. The proposed system is tested on the KITTI dataset and has shown an average accuracy of 2 meters in finding the final location of the vehicle.

## 1. INTRODUCTION

This paper introduces a visual localization method for autonomous vehicles AVs using input data from only one camera. Visual localization is studied for more than one decade, mostly in the fields of robotics and computer science (1). "Visual Localization" refers to finding the location by first taking pictures from the surrounding area (called query image), then, comparing those with a set of geotagged images (called database) to find the best matched images (2, 3), and finally, assigning a global location to the query images based on the location of their matches.

The ability to find the location is one of the basic requirements for any automated mobile robot. While the primary way for localization is to use Global Positioning System (GPS), introducing other methods seems necessary. This necessity arises from two main reasons. First, GPS may not work accurately in places like tunnels or among tall buildings where satellite signals cannot be received appropriately. Second, due to the probability of malfunctioning of the GPS-based navigation system, a backup method for localization must be developed if fully automated navigation is the ultimate goal.

The proposed method takes pictures of the surrounding area with a known frequency as the input and finds the location of the vehicle (camera) as accurately as possible as the output. The proposed method uses a feature-based algorithm to find the best match among a database of geotagged images. The geotag of the matched image serves as an approximate location for the query image. Besides, the displacement of the vehicle during a time interval can be computed given an estimate of vehicle speed. Combining this information using the Kalman filter, the location of the query images can be estimated more accurately.

The remainder of this paper is organized as follows. Section 2 presents the current literature. Section 3 provides an overview of the steps must be taken toward visual localization along with elaboration on specific algorithms that are used to solve the problem. The result of implementing the introduced method is presented in section 4. The paper ends with some final remarks in section 5.

---

[1] szahedi1@umd.edu

[2] kfarokhi@umd.edu

[3] amirn@umd.edu

## 2. LITERATURE

As it is mentioned earlier, in the literature of mobile robot navigation, the problem of visual localization is studied for almost two decades (4). The first part of the problem is image retrieval, for which several algorithms have been developed (5, 6, 7). Generally, there are two categories of image retrieval methods. The first category, **2D image-based localization**, tries to find the location of the query image through finding its similar ones in the database (8, 9). The accuracy of these methods increases when more than one image is retrieved and used for localization (10, 11). The second category, **3D structure-based localization**, uses the 3D model of the scene obtained from SfM (Structure from Motion) as its database. Matching the 2D query image with the 3D model, the camera orientation and position can be computed (12, 13).

One of the approaches traditionally used to find correspondent images in either of the two image retrieval categories is using local feature matching algorithms. This type of algorithms tries to find essential points of an image (called keypoint) and describe them by a vector of features (called descriptor). SIFT (14), SURF (15), FAST (16) and their variants (17,18) are the most popular algorithms of this kind each of which extracts the keypoints of the query image hence its descriptors. Comparing descriptors of the query image with those of database images based on a similarity measurement, images with the greatest number of correspondent keypoints are considered as best matches for the query image. Another approach to visual localization problem is to use Machine Learning (ML) which tries to find the location using global features of a scene (19, 20). One of the advantages of these methods is that they can find the location even when the photo belongs to an area off from city limits (20). Combining ML-based methods with local feature-based methods while searching for the best match has also revealed significant improvement in terms of efficiency (21, 22).

The output of the above methods is an approximate location for the query image which is not accurate enough for operation of AVs. Perspective-n-Points (PnP) algorithms are widely used to find the exact six degrees of freedom (6-DoF) query camera pose (23, 24). For PnP algorithms to obtain an exact camera pose, a 3D model of the scene is often required. A different approach to finding a more accurate location is to use statistical and filter-based methods. These types of methods aim to lower the cost of localization in terms of required hardware and time efficiency. In fact, instead of acquiring the exact solution, these methods try to estimate uncertainty while accepting such uncertainty up to a specific level to lower the costs associated with both implementation and application (25). In this regard, utilizing estimation methods such as Kalman Filter (26) has shown to provide acceptable results in real-time applications (27).

Studies who focus on AVs localization merely use a combination of methods introduced above to construct a complete system (28, 29, 30). The shortcoming of these systems is that they rely on advanced technologies or demand for complicated computational steps that increase the system run time. This paper introduces a system that requires the AV to be equipped with only one camera, rather than any complicated sensor. This system uses a statistical method of Kalman filter to improve the accuracy and decrease the uncertainty of the final localization.

## 3. METHODOLOGY

As it was mentioned before, this paper provides a step by step localization applicable for AVs. It assumes that a database of geotagged images extracted from a continuous video recording is available. Besides, it assumes that the AV is equipped with at least one camera that can take pictures of the surrounding area in predetermined time steps. Two major steps of the proposed localization method are image retrieval and location estimation using the Kalman filter. These two steps are explained with more details in the following subsections. To have a better understanding, the architecture of the proposed system is demonstrated in Figure 1.

### 3.1. Image retrieval from database of geotagged images

Image retrieval is comparing an image with a database to extract those images that plot almost the same scene. In this context, the image to be compared is the query image taken by a camera installed in a vehicle, and the database contains images that are previously taken from urban areas and tagged with their location. Consequently, the retrieving query image from the database provides an approximate location for the vehicle. This section explains how we use image retrieval to find an initial vehicle location during predetermined time intervals.

Generally, each image is an *n*-dimensional vector of features represented by numbers. However, there are some points within an image that are locally distinctive. These points are suitable candidates for comparing two images. One of the most popular local feature extractor algorithms is the Scale-Invariant Feature Transform (SIFT) algorithm which finds locally distinctive points of an image (14). SIFT algorithm has two main outputs:

- Keypoints which contain the pixel coordinates of interest points.
- Descriptors which are 128-dimensional arrays describing the local surrounding of each keypoint.

After running the SIFT algorithm for each image, we have a list of keypoints and their descriptors. Since these points are selected in a way that they are invariant to scale, rotation, and illumination, they are good candidates for comparing two images taken from the same scene in different times with different cameras. This means that instead of saving an image, its keypoints and their associated descriptors can be saved. Besides, the SIFT algorithm can be applied to each query image to extract their keypoints as well. Now, for each query image, we should look for an image within the database that shares the most common keypoints, called correspondences.

Here, we need some objective criteria to determine when two keypoints can be considered as correspondences. When comparing two keypoints, we are indeed comparing their 128-dimensional descriptors. There are two measures of effectiveness that are proved to be useful for comparing two keypoints (*10, 14*). The first criterion is shown in equation (1) states that for a keypoint from query image to be matched with a keypoint in a database image, the ratio distance between its closest and second closest match must be less than a predetermined threshold $\tau_1$.

$$\frac{d^2(g, f_{1^{st}})}{d^2(g, f_{2^{nd}})} < \tau_1^2 \qquad (1)$$

where, $g$ is the descriptor of the query keypoint, and $f_{1^{st}}$ and $f_{2^{nd}}$ are descriptors of its closest and second closest matches (correspondences) within the database images. The second criterion is the well-known cosine similarity measurement. Equation (2) states that two keypoints can be considered as correspondences if their associated cosine similarity is more than some threshold $\tau_2$.

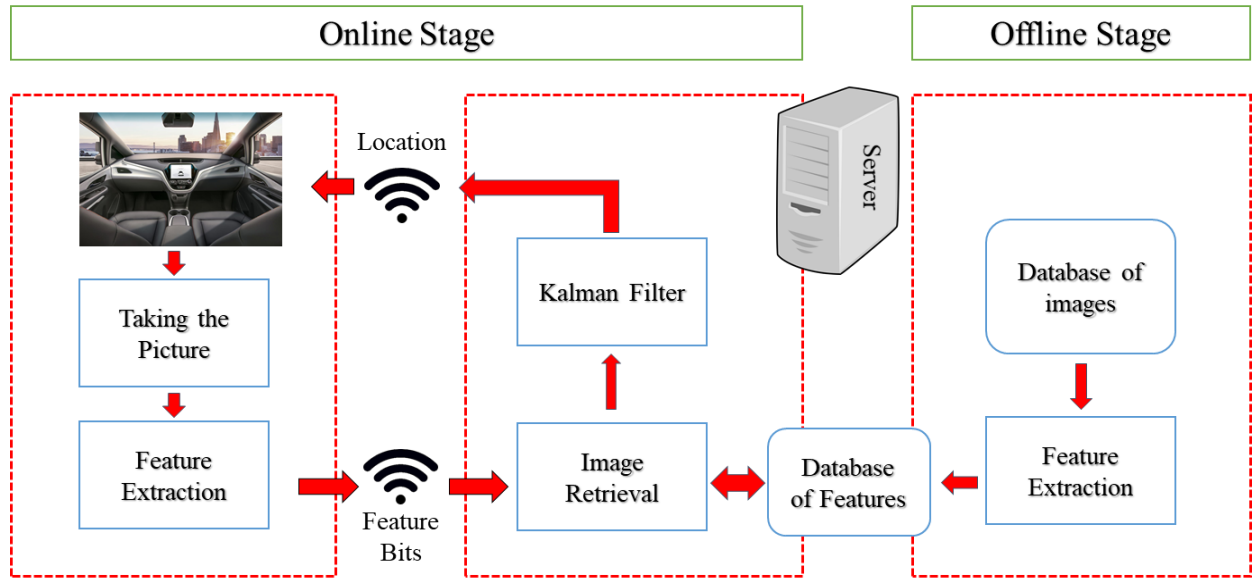

**Figure 1 Architecture of the proposed visual localization system.**

$$\cos(g, f) = \frac{f^T g}{\|f\|\|g\|} > \tau_2 \tag{2}$$

where $\|f\|$ and $\|g\|$ are length of $f$ and $g$ in Euclidian space.

After finding the correspondences between the query image and each image of the database, the best match is merely the image that has the most number of correspondences. Although the process of finding the best match based on this method is straightforward and easy to implement, it can be inefficient if we loop through an extensive database. However, in the case of AVs, the database can be effectively diminished by taking advantage of the last available GPS location or utilizing Geolocalization methods (20). Besides, instead of simple looping through the database, one can use advanced search algorithms that have a shorter running time (31).

### 3.2. Utilizing Kalman filter to find a more accurate location

The result of the first step is the approximate location for the car during specific time intervals. Given that query images were taken during a short time, the vehicle can be considered as a moving object with constant speed during that time. The next step is combining this information to find a better estimation of vehicle location at the next time step.

Our method uses the Kalman Filter to increase the accuracy of localization over time. Kalman filter, also known as linear quadratic estimation (LQE), is a popular stochastic method with various applications in autonomous navigation (*32*). In general, Kalman filter is an optimal estimator that infers the state of interest of a system from indirect, uncertain and inaccurate observations using joint probability distributions. Given that a discrete-time system has a state of $X_k \in R^n$ governed by the equation (3) and measurement of $y_k \in R^m$ expressed by equation (4) at the time step $k$, Kalman filter estimates $X_k$ in a way that it minimizes the mean square error.

$$X_k = F_k X_{k-1} + B_k u_k + w_{k-1} \tag{3}$$

$$y_k = H_k X_k + v_k \tag{4}$$

where $F_k$ is the state transition model, $B_k$ is the control model applied to control vector $u_k$ (optional), $H_k$ is the observation model, and $w_k$ and $v_k$ are process and measurement noises, respectively. An underlying assumption of Kalman filter is that $w_k$ and $v_k$ are independent and white normally distributed variables, i.e., $w_k \sim N(0, Q_k)$ and $v_k \sim N(0, R_k)$. Kalman filter algorithm includes two steps of prediction and update. In a simple language, it uses the dynamic model at time $k-1$ to predict the state at time $k$ and then uses the measurement at time $k$ to update the state.

In the case of our problem, where we aim to estimate the location of a vehicle by measuring its location in a relatively short distance, the model can be as easy as the model of a moving object with constant velocity. With this assumption, and in the absence of any measurement of the vehicle movement, system state vector, state transition, and observation models can be defined as equations (5) to (7).

$$X = \begin{bmatrix} x \\ y \\ \dot{x} \\ \dot{y} \end{bmatrix} \tag{5}$$

$$F = \begin{bmatrix} 1 & 0 & \Delta t & 0 \\ 0 & 1 & 0 & \Delta t \\ 0 & 0 & 1 & 0 \\ 0 & 0 & 0 & 1 \end{bmatrix} \quad (6)$$

$$H = \begin{bmatrix} 1 & 0 & 0 & 0 \\ 0 & 1 & 0 & 0 \end{bmatrix} \quad (7)$$

where $x,y$ are location coordinates, $\dot{x}$ is velocity in $x$ direction, and $\dot{y}$ is velocity in $y$ direction. $\Delta t$ is time distance between consecutive measurements. This means that in each time step we only have a measurement of the vehicle location (obtained from image retrieval), and we assume that the vehicle is moving with the constant speed. Although we usually have more information about the vehicle state, the goal of this paper is to show how we can get acceptable accuracy in localization with the minimum available information. Next section aims to demonstrate this by applying the method to a dataset.

## 4. RESULTS

As it was described in the previous sections, our localization system includes two straightforward steps of image retrieval and location estimation using Kalman filter. This section provides results of applying the introduced localization procedure using dataset introduced in the previous section. Before presenting any visual or numerical results, it is necessary to define system parameters. Table 1 shows the value of essential parameters that are used in this example. Additionally, Figure 2 shows a sample database image in which keypoints are extracted.

In this study, we use KITTI dataset to test our proposed system. KITTI dataset that is prepared and designed for robotics and autonomous driving researches contains records of traffic through color, and grayscale cameras, a Velodyne 3D laser scanner and a high-precision GPS/IMU inertial navigation system all installed on the top of a Volks Wagon station wagon (33). In this paper, it is assumed that the only data available to the vehicle is a database of geotagged images. The dataset contains photos recorded by four cameras, two grayscale cameras, and two color cameras, two on each side. There exists a total of almost 8500 photos, ten photos per second, for each camera. To build the desired dataset for the use in this paper, we picked the images taken from the right grayscale camera and associated them with their timestamp and GPS location of that timestamp. Therefore, our database is geotagged images whose timestamps serve as their label. Another data preparation step that improves the image retrieval efficiency is to apply the SIFT algorithm to all database images and save each image keypoints and descriptors in advance. Since our image retrieval only needs keypoints' information, there is no need to keep the pictures except their keypoints descriptors.

According to our data preparation steps, instead of images themselves, their keypoints and descriptors are saved. Thus, each image in the database is a *(k,*128*)* array, where *k* is the number of keypoints in that image. Now that the database is ready, we can apply our method to find the location of the moving vehicle at time *t*. To do that, we assume that the vehicle's camera starts taking pictures at time $t-5$ and takes six pictures, one each second. Here, we randomly choose time *t*, and it is easy to find desired query images taken from the left camera as timestamps of images are known.

**Table 1 Value of the parameters used in different steps of introduced method.**

| Parameter | Explanation | Value |
|---|---|---|
| $\tau_1$ | Distance ratio threshold in keypoint matching | 0.8 |
| $\tau_2$ | Cosine similarity threshold in keypoint matching | 0.97 |
| R | Location measurement noise that is considered to be constant | 0.0001 (in decimal degree) |
| Q | Initial State prediction noise which is a 4×4 matrix for our problem | I × 1000 |
| $\Delta t$ | Query images time interval | 1 sec |

The next step is to find the best match for each query image. To do so, we first apply the SIFT algorithm to all query images and find their keypoints. Then for the first query image, we loop through the whole database and by comparing descriptors and counting the number of correspondences according to the method explained in section 3.1, we find the best match for the first query image. For the following query images, instead of looking through the whole database, we look for the best match within a ±20 seconds time window around the best match of the first query image in the database. By doing so, image retrieval time reduces significantly. Figure 3 shows an example of two matched images with their corresponding keypoints connected with lines. To get a numerical sense, Table 2 shows the output of the image retrieval step for one series of query images. Note that in this table we also present the ground truth locations of the query images.

Geotags of the best matches are the input of the next step of the localization. In fact, these are the measurements of the vehicle's location in each second. Kalman filter uses these inaccurate measurements and tries to provide a better estimation of the location assuming constant speed for the vehicle. Table 3 is presented to show the location estimation of the Kalman filter for the same query image used in Table 2. In this table measurement and estimation, inaccuracies are presented in meters (using Haversine formula (34)) to show how Kalman filter improves the estimate in each time step.

Table 3 provides useful information about both steps of our method. According to this table, measurement noise tends to be around 18 meters which is roughly the distance that the database vehicle moves in 1 second with an average speed of 65 km/h. To reiterate for each query image while looking for the best match, we ignored images that are in one second distance from the query image. The reason for that was to simulate a more realistic situation. Therefore, the fact that the best matches are almost always in one second distance from their correspondence query image shows that image retrieval step can still find the best possible match even though an artificial restriction is imposed on the database.

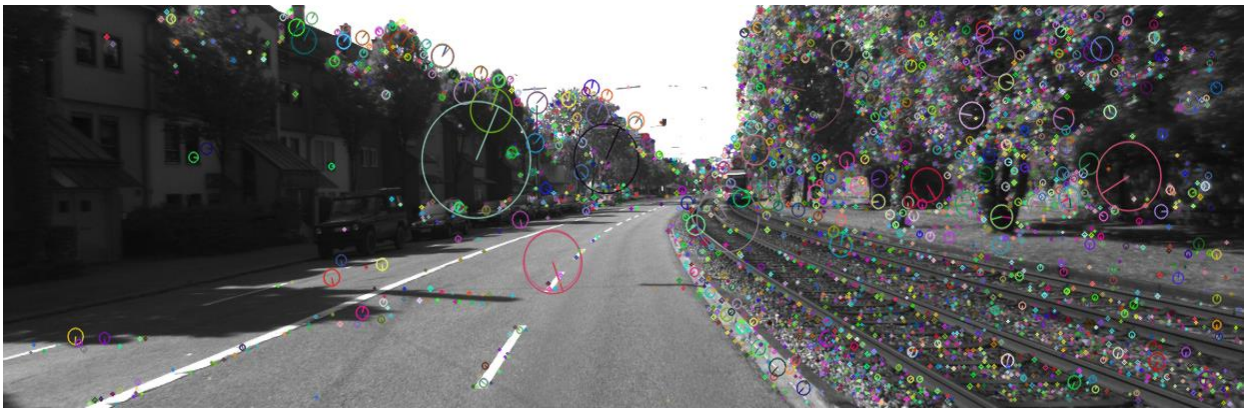

**Figure 2 A database image including 3200 keypoints.**

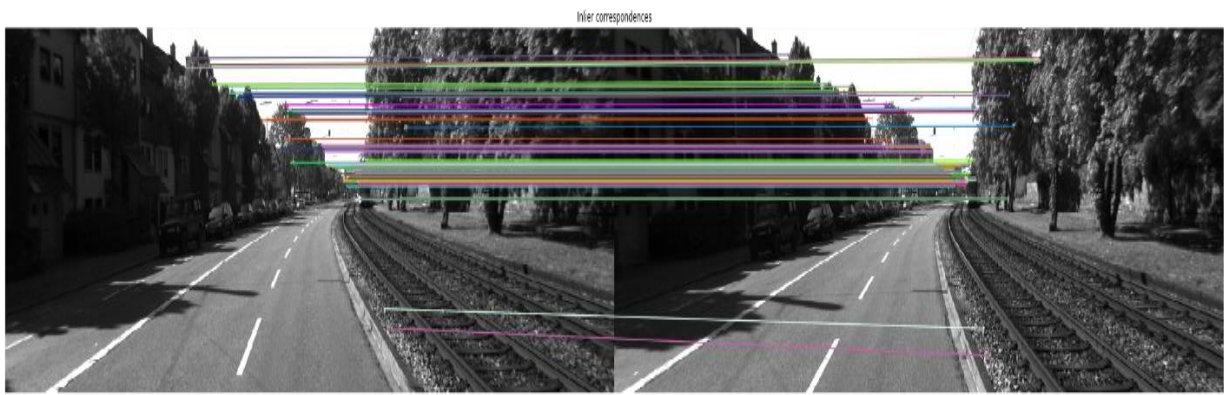

**Figure 3 An example of two matched image with their correspondences (left image is the query image and the right one is its best match from database).**

This example illustrates how Kalman filter performance improves in each time step. In the first two steps, the estimation of the Kalman filter is worse than our measurement. For the first step, the estimate is too close to the measurement that is understandable as we initially set the estimation uncertainty to be much higher than measurement noise. In the second time-step, our measurement is not only inaccurate but also in the opposite direction of the vehicle movement. In fact, the best-matched image has been detected to be the image taken one second before. In this step, the Kalman filter still has an uncertain model, and its input measurement is also in the opposite direction. These together cause the estimation to be much worse than measurement. However, in the following steps, the Kalman filter can use speed information and improve its estimation to correct the measurement more accurately.

Figure 4 illustrates the average measurement and estimation error of applying the proposed framework on a variety of query images for each time steps. As it can be seen in this figure the average final location accuracy of 1.85 meters is achieved. Exploring different examples reveals that the Kalman filter estimation and thus ultimate localization accuracy can be improved by retrieving those database images that are in line with the moving direction of the vehicle.

## 5. SUMMARY AND CONCLUSION

In this paper, we introduced a visual-based localization method including two main steps to solve AVs locating problem where no accurate GPS signal is available. Assuming a situation where a vehicle loses its GPS connection, image-based localization helps it to stay on track. Intuitively, the first step in this procedure is to find an approximate location by gathering information about the surrounding area. To do so, we proposed an image retrieval algorithm that takes consecutive query images as input and tries to find the approximate location for each image by finding images in the database that resemble them the most. The image retrieval algorithm employed in this study uses SIFT local feature descriptors for image matching. Once approximate location of the vehicle in consecutive time steps are obtained, a statistical method like Kalman filter can be used to combine this information with the general information about the vehicle movement and estimate a more accurate location for the vehicle. Here, we used a basic Kalman filter that assumes constant speed for the car and estimates the location in each time step. Combining these two steps, a localization system is created that is implemented in Python to be tested on the KITTI dataset. This dataset includes photos taken by different cameras installed on the top of a moving vehicle along with various information about the vehicle movement and location.

**Table 2 An example of the output image retrieval step.**

| Time | | $t-5$ | $t-4$ | $t-3$ | $t-2$ | $t-1$ | $t$ |
|---|---|---|---|---|---|---|---|
| Geotag of the best match | Longitude | 8.43413 | 8.43429 | 8.43413 | 8.43398 | 8.43384 | 8.43369 |
| | Latitude | 49.01494 | 49.01500 | 49.01494 | 49.01489 | 49.01484 | 49.01479 |
| Actual GPS coordinates | Longitude | 8.43429 | 8.43413 | 8.43398 | 8.43384 | 8.43369 | 8.43355 |
| | Latitude | 49.01500 | 49.01494 | 49.01489 | 49.01484 | 49.01479 | 49.01475 |

**Table 3 Kalman filter estimation of vehicle location in each time step.**

| Time | | $t-5$ | $t-4$ | $t-3$ | $t-2$ | $t-1$ | $t$ |
|---|---|---|---|---|---|---|---|
| Kalman filter estimates | Longitude | 8.43412 | 8.43445 | 8.43408 | 8.43387 | 8.43371 | 8.43355 |
| | Latitude | 49.01493 | 49.01506 | 49.01492 | 49.01485 | 49.01479 | 49.01474 |
| Measurement inaccuracy (in meters) | | 18.99 | 18.99 | 17.57 | 16.52 | 17.57 | 16.19 |
| Estimation inaccuracy (in meters) | | 19.81 | 38.77 | 11.52 | 3.28 | 2.31 | 0.94 |

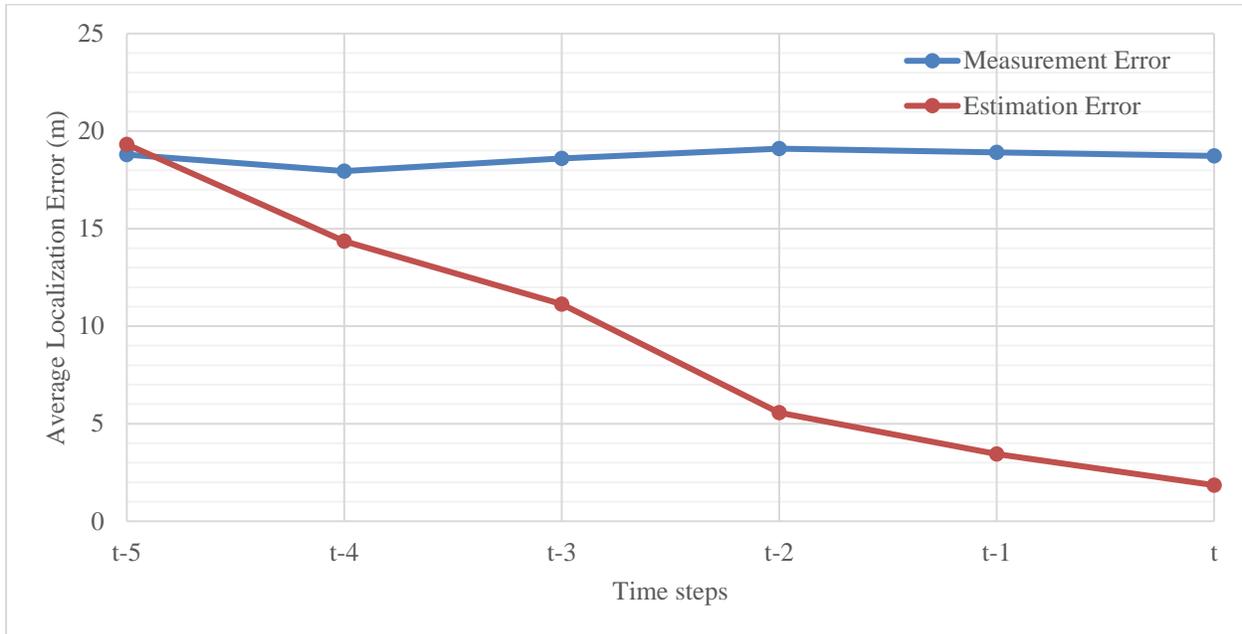

**Figure 4 Average measurement and estimation error of applying the framework on a variety of query images**

Applying our method to a version of KITTI dataset designed in a way that applies to our problem, we observed that accuracy as high as 2 meters is achievable using the simple two-step method introduced here. Although the technique here is not practical in its current form, it is useful as it shows how much accuracy one can expect from visual localization in the absence of any equipment except a simple camera and connection to a database of the geotagged images. We believe that the proposed method while following its general idea, can be modified and be improved to provide a relatively accurate location for AVs in real time. In addition to the fact that these types of methods are a good alternative under the situation where GPS readings are not reliable, they can be used in parallel with GPS as they can provide a more accurate estimation of the vehicles' location.

For future works, there are two types of improvements that can be added to the current basic method. Firstly, the accuracy of the final locating can be improved by considering the relative pose of the cameras of the database and query images. Having more information about the cameras and associating the database with the location of some points in the scene rather than the camera location, this relative pose can be easily obtained using spatial geometry. Secondly, one can assume there exist more information about the vehicle at each time, for example, its speed, acceleration, odometry and so on. This measurement may be used as input to an Extended Kalman Filter (EKF) which can provide a more accurate estimation of location. For the whole system to be efficient and have a real-time application, more effort must be put on the image retrieval step. For instance, while preparing the dataset, instead of saving all the keypoints detected by the SIFT algorithm, only those keypoints that are more likely to be found in other images can be saved. Detecting these keypoints that can be considered as consistent inliers requires information from different matches for each database image. A practical way to do that is to match each database image with its neighbors in specific time distance and find inliers using a sampling algorithm. These inliers may be weighted based on their consistency in different matches.

**REFERENCES**

1. Se, S., D. Lowe, and J. Little. Mobile robot localization and mapping with uncertainty using scale-invariant visual landmarks. *The international Journal of robotics Research*, 2002. 21: 735-758.
2. Arandjelović, R., and A. Zisserman. Scalable descriptor distinctiveness for location recognition. *Asian Conference on Computer Vision*, 2014. 188-204.
3. Torii, A., J. Sivic, T. Pajdla, and M. Okutomi. Visual place recognition with repetitive structures. *IEEE conference on computer vision and pattern recognition*, 2013. 883-890.
4. Lowry, S., N. Sünderhauf, P. Newman, J. J. Leonard, D. Cox, P. Corke, and M. J. Milford. Visual place recognition: A survey. *IEEE Transactions on Robotics*, 2016. 32: 1-19.
5. Chum, O., J. Philbin, J. Sivic, M. Isard, and A. Zisserman. Total recall: Automatic query expansion with a generative feature model for object retrieval. *IEEE 11th International Conference*, 2007, October. 1-8.
6. Jegou, H., H. Harzallah, and C. Schmid. A contextual dissimilarity measure for accurate and efficient image search. *IEEE, Computer Vision and Pattern Recognition*, 2007. 1-8.
7. Philbin, J., O. Chum, M. Isard, J. Sivic, and A. Zisserman. Object retrieval with large vocabularies and fast spatial matching. *Computer Vision and Pattern Recognition, CVPR'07. IEEE Conference on*, 2007, June. 1-8.
8. Chen, D., S. Tsai, V. Chandrasekhar, G. Takacs, H. Chen, R. Vedantham, R. Grzeszczuk, and B. Girod. *Signals, Systems and Computers (ASILOMAR), 2011 Conference Record of the Forty Fifth Asilomar Conference on*. IEEE, 2011.
9. Gronat, P., G. Obozinski, J. Sivic, and T. Pajdla. *Proceedings of the IEEE conference on computer vision and pattern recognition.* 2013. 907-914
10. Zhang, W., and J. Kosecka. Image Based Localization in Urban Environments. *In 3DPVT,* 2006. 6: 33-40.
11. Sattler, T., A. Torii, J. Sivic, M. Pollefeys, H. Taira, M. Okutomi, and T. Pajdla. Are Large-Scale 3D Models Really Necessary for Accurate Visual Localization? *2017 IEEE Conference on Computer Vision and Pattern Recognition (CVPR).* IEEE, 2017. 6175-6184.
12. Li, Y., N. Snavely, D. Huttenlocher, and P. Fua. Worldwide pose estimation using 3d point clouds. *European conference on computer vision,* 2012. 15-29.
13. Kukelova, Z., M. Bujnak, and T. Pajdla. Real-time solution to the absolute pose problem with unknown radial distortion and focal length. *Proceedings of the IEEE International Conference on Computer Vision.* 2013. 2816-2823.
14. Lowe, D. G. Distinctive image features from scale-invariant keypoints. *International journal of computer vision*, 2004. 60(2): 91-110.
15. Bay, H., T. Tuytelaars, and L. Van Gool. Surf: Speeded up robust features. *European conference on computer vision,* 2006. 404-417.
16. Rosten, E., and T. Drummond. Machine learning for high-speed corner detection. *European conference on computer vision,* 2006. 430-443.
17. Rosten, E., R. Porter, and T. Drummond. Faster and better: A machine learning approach to corner detection. *IEEE transactions on pattern analysis and machine intelligence*, 2010. 32(1): 105-119.

18. Rublee, E., V. Rabaud, K. Konolige, and G. Bradski. 2011, November. ORB: An efficient alternative to SIFT or SURF. *Computer Vision (ICCV), 2011 IEEE international conference on,* 2011. IEEE. 2564-2571.
19. Weyand, T., I. Kostrikov, and J. Philbin. Planet-photo geolocation with convolutional neural networks. *European Conference on Computer Vision*, 2016. 37-55.
20. Hays, J., and A. A. Efros. 2008, June. IM2GPS: estimating geographic information from a single image. *Computer Vision and Pattern Recognition, 2008. CVPR 2008. IEEE Conference on,* 2008. IEEE. 1-8.
21. Zhang, S., M. Yang, T. Cour, K. Yu, and D. N. Metaxas. Query specific rank fusion for image retrieval. *IEEE Transactions on Pattern Analysis and Machine Intelligence*, 2015. 37(4): 803-815.
22. Zamir, A.R. and M. Shah. Image geo-localization based on multiple nearest neighbor feature matching using generalized graphs. *IEEE transactions on pattern analysis and machine intelligence,* 2014. 36(8): 1546-1558.
23. Li, Y., N. Snavely, and D. P. Huttenlocher. Location recognition using prioritized feature matching. *European conference on computer vision,* 2010. 791-804.
24. Li, Y., N. Snavely, D. P. Huttenlocher, and P. Fua. Worldwide pose estimation using 3d point clouds. *European conference on computer vision,* 2012. 15-29.
25. Smith, R., M. Self, and P. Cheeseman. Estimating uncertain spatial relationships in robotics. *Autonomous robot vehicles,* 1990. 167-193.
26. Kalman, R.E. A new approach to linear filtering and prediction problems. *Journal of basic Engineering,* 1960. 82(1): 35-45.
27. Wong, X. I., and M. Majji. Extended Kalman Filter for Stereo Vision-Based Localization and Mapping Applications. *Journal of Dynamic Systems, Measurement, and Control,* 2018. 140(3): 030908.
28. Wolcott R. W., and R. M. Eustice. Visual localization within lidar maps for automated urban driving. *Intelligent Robots and Systems (IROS 2014), 2014 IEEE/RSJ International Conference on*, 2014. 176-183. IEEE.
29. Trepagnier P. G., J. E. Nagel, P. M. Kinney, M. T. Dooner, B. M. Wilson, C. R. Schneider Jr, and K. B. Goeller, inventors; Gray and Co Inc, assignee. Navigation and control system for autonomous vehicles. United States patent US 8,050,863. 2011 Nov 1.
30. Trepagnier P. G., J. E. Nagel, P. M. Kinney, M. T. Dooner, B. M. Wilson, C. R. Schneider Jr, and K. B. Goeller, inventors; Gray and Co Inc, assignee. Navigation and control system for autonomous vehicles. United States patent US 8,346,480. 2013 Jan 1.
31. Wu, X., and K. Kashino. Second-order configuration of local features for geometrically stable image matching and retrieval. *IEEE Transactions on Circuits and Systems for Video Technology,* 2015. 25(8): 1395-1408.
32. Laaraiedh, M. Implementation of Kalman filter with python language. *arXiv preprint arXiv:1204.0375,* 2012.
33. Geiger, A., P. Lenz, C. Stiller, and R. Urtasun. Vision meets robotics: The KITTI dataset. *The International Journal of Robotics Research,* 2013. 32(11): 1231-1237.
34. Robusto, C. C. The cosine-haversine formula. *The American Mathematical Monthly*, 1957, Jan 1, pp. 38-40.